\documentclass{article}
\usepackage{ijcai22}

\usepackage{times}
\usepackage{soul}
\usepackage{url}
\usepackage[hidelinks]{hyperref}
\usepackage[utf8]{inputenc}
\usepackage[small]{caption}
\usepackage{graphicx}
\usepackage{amsmath}
\usepackage{amsthm}
\usepackage{booktabs}
\usepackage{algorithm}
\usepackage{algorithmic}
\usepackage{xcolor}
\usepackage{subcaption}
\usepackage{standalone}

\usepackage{pgfplots}
\usepackage{pgfplotstable}
\pgfplotsset{compat=1.17}  

\title{Probe-Based Interventions for Modifying Agent Behavior}

\author{
Mycal Tucker$^1$
\and
William Kuhl$^1$\and
Khizer Shahid$^1$\and
Seth Karten$^2$\and
Katia Sycara$^2$\and
Julie Shah$^1$
\affiliations
$^1$Massachusetts Institute of Technology\\
$^2$Carnegie Mellon University\\
\emails
\{mycal, julie\_a\_shah\}@csail.mit.edu,
\{bkuhl, khizer\}@mit.edu,
\{skarten, katia\}@cs.cmu.edu
}

\begin{document}

\maketitle

\begin{abstract}
Neural nets are powerful function approximators, but the behavior of a given neural net, once trained, cannot be easily modified.
We wish, however, for people to be able to influence neural agents' actions despite the agents never training with humans, which we formalize as a human-assisted decision-making problem.
Inspired by prior art initially developed for model explainability, we develop a method for updating representations in pre-trained neural nets according to externally-specified properties.
In experiments, we show how our method may be used to improve human-agent team performance for a variety of neural networks from image classifiers to agents in multi-agent reinforcement learning settings.
\end{abstract}

\section{Introduction}
We envision a world in which autonomous agents learn patterns from data on their own but, when given human guidance, they respond appropriately.
Such agents would enable high individual and team performance, as well as bridge the gap between data-driven and symbolic reasoning.

Neural nets are a natural architectural choice for such agents.
As universal function approximators, with enough data, they can learn correlations to map from inputs to desired outputs.
As a result, they have been applied in a wide range of contexts, including image recognition, self-driving cars, and communication domains \cite{classifier,ic3net}.
Any single neural net is trained to map from its inputs (e.g., sensor data from a self-driving car) to outputs (e.g., throttle control).

While neural networks generally are quite flexible, any given neural net, once trained, acts as a black box that maps from inputs to outputs.
For example, a neural net on a self-driving car that normally accepts images cannot accept verbal commands at test time.
However, we wish for agents to act more like humans, who can change their behaviors based on new sources of information.
The challenge is to do so with pre-trained neural net agents that have never been trained to accept the new form of input.

\begin{figure}[thb]
    \centering
    \includegraphics[trim={0cm 0cm 0cm 0.0cm}, clip=true, width=0.47\textwidth]{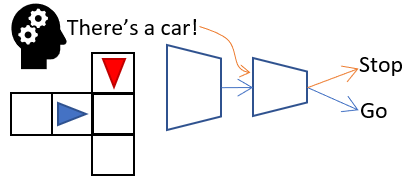}
    \caption{If a neural agent (blue, left) fails to see a nearby car (red, top), it may act unsafely. We enable humans to inject information into representations, modifying behavior without fully dictating it.}
    \label{fig:intro_diagrma}
\end{figure}

In this work, the key insight is that, if a neural net already models some concept in its representation space (e.g., ``there is another car nearby''), the problem of ``injecting'' information becomes one of updating that symbolic representation.
While neural representation spaces may be difficult for humans to understand, one may train separate neural net ``probes'' to map from representations to symbolic properties \cite{alain2016understanding,whatif}.
Lastly, one may use these probes to update representations according to a desired property.
For example, as depicted in Figure~\ref{fig:intro_diagrma}, one may train a probe to predict, from a representation, if there is another car nearby, and then use that probe to update the representation to reflect the fact that there actually is a car that the neural net failed to notice.

In this paper, we make three contributions.
First, we define the problem of ``human-assisted decision-making'' to formalize an intuitive desire of leveraging both human and agent observations in decision-making settings.
Next, we describe how techniques that had earlier been developed for neural net explanations may be re-purposed to solve our problem.
Third, we propose an improvement over these techniques that mitigates a limitation of prior art, based on a theory of redundant encodings.
In experiments, we show how our method can be used to improve the performance of an image classifier (without retraining the classifier), prevent crashes in a simulated road intersection environment in which cars fail to see each other, or enabled flawed agents in a collaborative grid-world environment to find each other.

\section{Related Work}

\subsection{Interpretability and Explainability}
Interpretability and explainability techniques focus on developing models that are easily understandable, or techniques to better understand pre-trained models \cite{molnar2019}.

On the one hand, through special architectural or training choices, one may train agents with interpretable representation or decision-making systems.
For example, in models trained with concept-whitening, it is easy to identify which neuron activations correspond to which concepts \cite{chen2020concept}.
More generally, techniques that shape representation spaces according to human design principles (e.g., prototype-based classification, disentanglement, etc.) can be used to support human understanding.
In some domains, like style-transfer, interepretable representations support mixing aspects of different representations to produce desired outputs \cite{john-etal-2019-disentangled}.

On the other hand, other research attempts to create human-understandable explanations of pre-trained neural networks with minimal assumptions about how those networks have been trained.
For example, post-hoc explanation techniques like LIME produce linear approximations of neural network decision boundaries \cite{ribeiro2016model}.
Other methods train simple classifiers to expose what information is present in neural representations, but not how such information is used \cite{alain2016understanding}.

These techniques are useful for understanding neural networks, but differ from our work in two ways.
First, unlike interpretability research, we do not assume that we can train the neural net model.
Second, we fundamentally seek to modify neural net behavior as opposed to only understanding it.

\subsection{Causal Analysis of Neural Networks}
Our work most directly builds upon work previously developed for causal analysis of neural models.
In causal analysis literature, researchers seek to identify how models use particular properties in their decision-making.
For example, an image classifier may often correctly classify images of dogs, but causal analysis may reveal that the model changes its predictions if a single input pixel changes.
This is an important causal property of the model.

Researchers have applied causal analysis techniques both on input-space features (e.g., if this pixel changes, how does the model prediction change) and latent features \cite{zeiler2014visualizing}.
We focus here on latent feature analysis.
One technique, quantitative testing of concept activation vectors (TCAV), measures how a neural network's predictions change when a representation is updated according to a conceptual property \cite{tcav}.
For example, the authors find a direction in the latent space that is correlated with images of striped objects, and they find that if they move representations along that direction, they increase the probability of the model predicting zebras.
In domains in which humans know what the ``right'' causal path for a particular output is, this provides a useful quality check.

More recently, researchers similarly used causal analysis techniques to study if large language models use syntactic properties in predictions.
While similarly motivated to TCAV - they wish to understand what latent properties models use in making predictions - they adopted more general frameworks.
Applying multiple linear classifiers appeared to induce better effects than the single classifier used by TCAV \cite{elazar2020amnesic}.
Furthermore, Tucker et al. \shortcite{whatif} developed a method to use non-linear classifiers to extract and modify syntactic properties.
Lastly, \cite{diagnostic} showed how one could update representations in LSTM-based language models and improve subject-verb agreement in sentences.

We focus on modifying agent behavior rather than just understanding how they act.
We implement the technique proposed by Tucker at al. \shortcite{whatif}, which encompasses TCAV as a special case, and use them as baselines for generating updated embeddings in non-language settings.
Lastly, we show how a new design further improves upon these baselines, enabling better control of agent behavior.

\section{Human-Assisted Decision Making Problem Formulation}
We define a new problem, ``human-assisted decision-making,'' based on the Decentralized Partially-Observable Markov Decision Process framework in which different agents only observe subsets of the true state \cite{bernstein2013complexity}.
At training time, $N$ artificial agents train to maximize some shared reward by learning policies ($\pi_1$ to $\pi_N$) that map from observations of the true state, $S$, to actions.
Reward is a function of the joint state and actions taken by all agents: $R(S, \pi_1(o_1), \pi_2(o_2), ...)$.

At test time, we define a human participant with an associated observation function, $O_h: S \rightarrow o_h$.
Human observations, an individual agent's policy, and the agent's observations, may be combined into a ``hybrid'' policy, $h: o_h, o_i, \pi_i \rightarrow a_i$.
That is, the hybrid policy uses the human's observation, the agent's observation, and the agent's policy to produce a new action.
This parallels human supervisory control of automated systems frameworks in which a human monitors an agent's output and decides when to intervene, but the roles of the human and agent are reversed.

The human-assisted decision-making problem is therefore one of generating a ``good'' hybrid policy, $h$, that reconciles human and agent observations within an agent policy.
The aim is to generate a hybrid policy that maximizes $R(S, h_1(o_h, o_1, \pi_1), \pi_2(o_2),..., \pi_N(o_N))$.
This paper frames the human-assisted decision-making problem in reinforcement learning contexts, but the framework may be adapted to other settings.

\section{Technical Approach}
We proposed probe-based interventions as an approach to the human-assisted decision-making problem.
Our method relies upon using human observations to update the internal representations of neural agent policies.


\subsection{Probe Design and Training}
For simplicity, we describe how to use probe-based interventions in feed-forward image classifiers, but our experiments illustrate the value of this approach for a variety of domains and neural architectures.

We characterize a feedforward neural network model, $M$, as composed of $N$ layers, which map from input image $x \in R^X$ to an output $y \in R^Y$.
We denote the representation generated after layer $k$ as $z_k \in R^Z$.
$M$ may be thought of as two neural networks: $z_k = M_{k-}(x)$ and $y = M_{k+}(z_k)$ where the subscripts for $M$ denote the indices of layers before or after layer $k$.

A probe is defined as a neural network mapping from representations generated at layer $k$ to predictions of a property, $s \in R^S$.
For example, a probe may predict, given a representation generated by an image classifier, if the image was originally of an animal or a vehicle.
The probe may be a linear classifier, but it may equally well be a non-linear, multi-layer neural network.

The probe may be trained using a dataset of $(z_k, s)$ pairs.
In TCAV, although not explicitly framed as training a neural net, a linear probe is trained with embeddings generated by supplying a dataset of input images describing some concepts, and those images are encoded to generate the corresponding $z_k$ \cite{tcav}.
In other works studying language models, non-linear probes are trained to map from representations of sentences in language models to aspects of the syntactic structure of the sentence \cite{whatif}.
Fundamentally, these probes may be trained just like any other neural network to map from representations to predictions.

\subsection{Counterfactual Generation}
Next, we show how to use such probes to update representations.
Adopting the method of \cite{whatif}, ``counterfactual'' representations, $z'$, are created by updating an original encoding, $z$, along the gradient of the loss function, $L$, computed on the output of the probe, $p$, and a desired property, $s'$.
That is, starting at $z' = z$, we incrementally update as follows, with step size $\delta$:

\begin{equation}
    z' = z' + \delta \nabla_{z'} L(p(z'), s')
\end{equation}

One may apply such updates until a stopping criterion, such as number of steps or minimum probe loss, is reached.
To then evaluate the effect of using counterfactual embeddings, one may pass $z'$ through the remainder of the model to generate a counterfactual outputs: $y' = M_{k+}(z')$.
If the original output, $y$, and $y'$ systematically differ, one may say that $M$ uses a representation that encodes information about $s$.

While this method is borrowed from Tucker et al.~\shortcite{whatif}, other techniques similarly use gradient updates, although only for linear classifiers.
TCAV measures, in the limit of small gradient steps, the change in model prediction with respect to the change in embedding.
Giulianelli~\shortcite{diagnostic} perform a single gradient update of fixed magnitude.

\begin{figure}
    \centering
    \includegraphics[trim={0cm 0cm 0cm 0.0cm}, clip=true, width=0.47\textwidth]{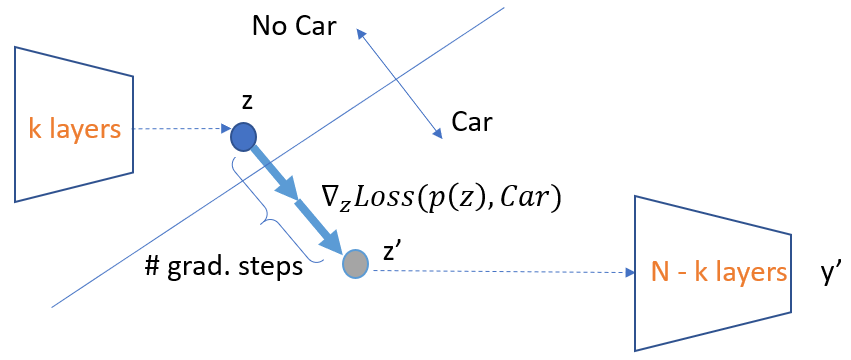}
    \caption{A graphical depiction of generating counterfactual embeddings. The first $k$ layers of an agent generate an embedding, $z$ (depicted here in 2D for simplicity). A counterfactual embedding, $z'$, is generated by moving in the embedding space along the gradient of the probe loss until the probe predicts ``Car.'' Here, the probe is depicted as a linear classifier, but it may be a multi-layer neural network. Lastly, $z'$ is passed through the rest of the agent to produce an action or decision.}
    \label{fig:updating}
\end{figure}

In this work, we use Tucker et al.'s \shortcite{whatif} incremental, gradient-based method for updating representations.
We note that, because this method depends upon gradients calculated through trained probes, different probe designs may produce different types of updates.

Fundamentally, this method for counterfactual generation is our proposed approach to the human-assisted decision-making problem.
We use probes to update the original representations generated by the agent to reflect new information supplied by the human, and we pass the resulting representation through the remainder of the agent's policy.

\subsection{Dropout Probes}
While the existing methods develop a gradient-based mechanism for creating counterfactual representations according to a trained probe, they could be susceptible to probe overfitting, which can produce undesirable counterfactual updates.

In particular, consider an image classifier that encodes a single property - is the input image an animal or a vehicle - in a redundant manner.
For example, if the representation vector is 32-dimensional but the model encodes the animal vs. vehicle property identically in two neurons, then a probe could arbitrarily learn to use either neuron in making its prediction.
Unfortunately, even if the model also used representations of animal vs. vehicle information, if the model used the other, redundant neuron, creating counterfacutal embeddings according to the probe would not affect model behavior.
Prior art has hinted at this possibility: TCAV, for example, instantiates many linear classifiers to see if, on average, they produce an effect, and Elazar et al. \shortcite{elazar2020amnesic} use an iterative approach to create multiple classifiers.
This can be thought of as a mechanism to ensure greater alignment between probe and model.

Instead of using a suite of classifiers, we instead propose the simple modification of adding a dropout layer before probes.
During training, the dropout layer randomly masks a subset of inputs with some probability $p$.
When generating counterfactual embeddings, dropout may be disabled.

With dropout, a probe must learn to use all subsets of representations that are informative of the predicted property, $s$.
Thus, we hypothesized that $\nabla_{z'} L(p(z'), s')$ (the gradient along which one updates $z'$), is less likely to be orthogonal to the model's decision boundary's gradient, and therefore that creating counterfactuals using dropout probes would change model behavior more than when using standard probes.

\section{Experiments}
We conducted three suites of experiments to evaluate: 1) the ability of probe-based interventions to affect neural net behavior and 2) the advantage of using dropout in probes.
First, we demonstrated how to boost an image classifier's performance be injecting symbolic information about high-level categories in the data.
Second, in a traffic-junction environment, we showed how probe-based interventions supplied by humans could prevent collisions even if cars didn't directly observe each other.
Lastly, in a multi-agent reinforcement learning environment, in which the agents needed to communicate to succeed at the task, we showed how our interventions enabled us to inject information into agents without needing to understand all aspects of the learned policies.

Experiments followed a standard procedure: 1) an agent was trained on the primary task (e.g., image classification) and then frozen, 2) a probe was trained on an auxiliary task to map from representations to a desired property, 3) at test time, we used the trained agents and probes to update embeddings and modify behavior.
Updated embeddings were generated using an SGD optimizer with learning rate 0.1 and momentum 0.9 for a fixed number of timesteps (specified in experiments) or until the probe loss was below 0.001.

In each experiment setup, we performed 5 trials by training new agents from scratch, and we studied the effect of dropout rate by training probes with different dropout rates, from 0 to 0.9 at increments of 0.1, for the same model.
Anonymized code for some experiments is available online; complete code will be edited for clarity and released upon acceptance.\footnote{\url{https://bit.ly/31xqbbf}}

\subsection{Image Classifier Boosting}
We showed that we could improve an image classifier's performance by injecting hierarchical information into representations, and that using dropout was an important part of boosting performance.

First, following a standard online tutorial for an image classifier, we trained a 6-layer convolutional neural net (CNN) on the CIFAR10 dataset, achieving roughly 50\% percent accuracy \cite{cifar10,classifier}.
Although this does not match state of the art classification performance on CIFAR, it follows standard practices and allows us to demonstrate how our method allowed use to improve model performance.

\begin{figure}[!t]
    \centering
    \begin{subfigure}[b]{0.48\textwidth}
        \centering
                \begin{tikzpicture}[scale=1.0] 
            \begin{axis}[legend pos=south east,  width={\textwidth},
                ylabel=Classification Accuracy,
    xlabel=Dropout Rate,
    xmin=-0.05, xmax=0.95,
    ymin=0.505,ymax=0.56,
    height=5.4cm,
      legend style={fill=none},
    title={CIFAR Classification Acc. Varying \# Steps}]
            \addplot [black!80]
                    table [x=x, y=y, y error=y-err]{
                        x	y y-err
                        0.0	0.51 0.017
                        0.1	0.51 0.017
                        0.2	0.51 0.017
                        0.3	0.51 0.017
                        0.4	0.51 0.017
                        0.5	0.51 0.017
                        0.6	0.51 0.017
                        0.7	0.51 0.017
                        0.8	0.51 0.017
                        0.9	0.51 0.017
                            };
            \addlegendentry{Baseline}
            \addplot [mark=o, blue!80, dashed, 
                      mark options={solid},
                      error bars/.cd, y dir=both, y explicit,
                      error bar style={line width=1pt,solid},
                      error mark options={line width=1pt,mark size=2pt,rotate=90}]
                    table [x=x, y=y, y error=y-err]{
                        x	y y-err
                        0.0	0.527 0.002
                        0.1	0.531 0.002
                        0.2	0.532 0.002
                        0.3	0.531 0.002
                        0.4	0.532 0.003
                        0.5	0.533 0.003
                        0.6	0.533 0.002
                        0.7	0.531 0.003
                        0.8	0.532 0.002
                        0.9	0.525 0.002
                            };
            \addlegendentry{$1$ layer; 200 steps}
            \addplot [mark=asterisk, violet!80, dashed, 
                      mark options={solid},
                      error bars/.cd, y dir=both, y explicit,
                      error bar style={line width=1pt,solid},
                      error mark options={line width=1pt,mark size=2pt,rotate=90}]
                    table [x=x, y=y, y error=y-err]{
                        x	y y-err
                        0.0	0.528 0.003
                        0.1	0.534 0.004
                        0.2	0.536 0.004
                        0.3	0.538 0.004
                        0.4	0.538 0.004
                        0.5	0.540 0.005
                        0.6	0.540 0.005
                        0.7	0.540 0.005
                        0.8	0.539 0.005
                        0.9	0.534 0.004
                            };
            \addlegendentry{$1$ layer; 400 steps}
            \addplot [mark=square, red!80, dashed, 
                      mark options={solid},
                      error bars/.cd, y dir=both, y explicit,
                      error bar style={line width=1pt,solid},
                      error mark options={line width=1pt,mark size=2pt,rotate=90}]
                    table [x=x, y=y, y error=y-err]{
                        x	y y-err
                        0.0	0.530 0.003
                        0.1	0.538 0.005
                        0.2	0.539 0.005
                        0.3	0.541 0.005
                        0.4	0.542 0.005
                        0.5	0.542 0.005
                        0.6	0.543 0.005
                        0.7	0.541 0.004
                        0.8	0.541 0.005
                        0.9	0.541 0.005
                            };
            \addlegendentry{$1$ layer; 600 steps}
            \end{axis} 
        \end{tikzpicture}
    \end{subfigure}
    \caption{Updating representations in the CIFAR10 classifier based on animal/vehicle class, using 1-layer probes, improved model accuracy. Interventions produced the greatest benefits when using high-dropout probes ($x$ axis) for many gradient steps (different lines).}
    \label{fig:cifar_dropout}
\end{figure}
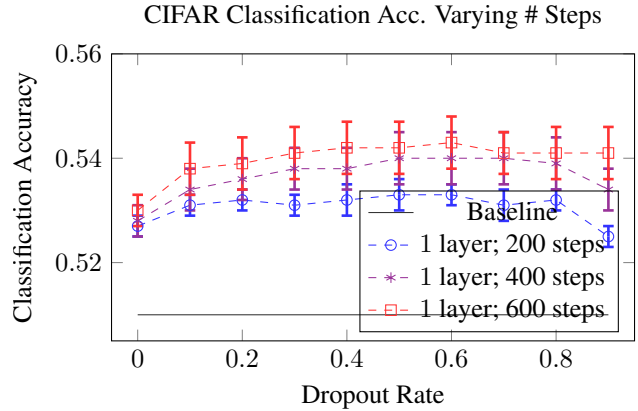

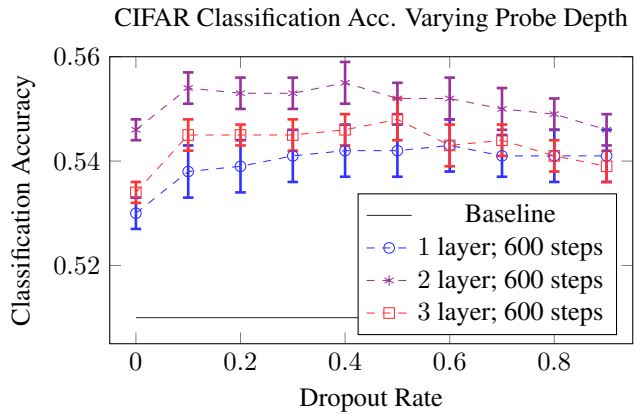
\begin{figure}[t!]
    \centering
    \begin{subfigure}[b]{0.48\textwidth}
        \centering
                \begin{tikzpicture}[scale=1.0] 
            \begin{axis}[legend pos=south east,  width={\textwidth},
                ylabel=Classification Accuracy,
    xlabel=Dropout Rate,
    xmin=-0.05, xmax=0.95,
    ymin=0.505,ymax=0.56,
    height=5.4cm,
    title={CIFAR Classification Acc. Varying Probe Depth}]
            \addplot [black!80]
                    table [x=x, y=y, y error=y-err]{
                        x	y y-err
                        0.0	0.51 0.017
                        0.1	0.51 0.017
                        0.2	0.51 0.017
                        0.3	0.51 0.017
                        0.4	0.51 0.017
                        0.5	0.51 0.017
                        0.6	0.51 0.017
                        0.7	0.51 0.017
                        0.8	0.51 0.017
                        0.9	0.51 0.017
                            };
            \addlegendentry{Baseline}
            \addplot [mark=o, blue!80, dashed, 
                      mark options={solid},
                      error bars/.cd, y dir=both, y explicit,
                      error bar style={line width=1pt,solid},
                      error mark options={line width=1pt,mark size=2pt,rotate=90}]
                    table [x=x, y=y, y error=y-err]{
                        x	y y-err
                        0.0	0.530 0.003
                        0.1	0.538 0.005
                        0.2	0.539 0.005
                        0.3	0.541 0.005
                        0.4	0.542 0.005
                        0.5	0.542 0.005
                        0.6	0.543 0.005
                        0.7	0.541 0.004
                        0.8	0.541 0.005
                        0.9	0.541 0.005
                            };
            \addlegendentry{$1$ layer; 600 steps}
            \addplot [mark=asterisk, violet!80, dashed, 
                      mark options={solid},
                      error bars/.cd, y dir=both, y explicit,
                      error bar style={line width=1pt,solid},
                      error mark options={line width=1pt,mark size=2pt,rotate=90}]
                    table [x=x, y=y, y error=y-err]{
                        x	y y-err
                        0.0	0.546 0.002
                        0.1	0.554 0.003
                        0.2	0.553 0.003
                        0.3	0.553 0.003
                        0.4	0.555 0.004
                        0.5	0.552 0.003
                        0.6	0.552 0.004
                        0.7	0.550 0.004
                        0.8	0.549 0.003
                        0.9	0.546 0.003
                            };
            \addlegendentry{$2$ layer; 600 steps}
            \addplot [mark=square, red!80, dashed, 
                      mark options={solid},
                      error bars/.cd, y dir=both, y explicit,
                      error bar style={line width=1pt,solid},
                      error mark options={line width=1pt,mark size=2pt,rotate=90}]
                    table [x=x, y=y, y error=y-err]{
                        x	y y-err
                        0.0	0.534 0.002
                        0.1	0.545 0.003
                        0.2	0.545 0.002
                        0.3	0.545 0.003
                        0.4	0.546 0.003
                        0.5	0.548 0.004
                        0.6	0.543 0.004
                        0.7	0.544 0.003
                        0.8	0.541 0.003
                        0.9	0.539 0.003
                            };
            \addlegendentry{$3$ layer; 600 steps}

            \end{axis} 
        \end{tikzpicture}
    \end{subfigure}
    \caption{Using multi-layer dropout probes for interventions in the CIFAR10 classifier produced greater benefits than linear probes. As before, introducing dropout also boosted performance.}
    \label{fig:cifar_layers}
\end{figure}

Probes were trained on a binary classification task: given representations taken from the third layer of the image classifier, the probe was trained to predict if the image was one of an animal or a vehicle.
At test time, we sought to improve the main image classifier's accuracy by injecting the ``animal vs. vehicle'' information into the representations.
We compared the effect of using probes with different numbers of layers and different dropout rates.
Our results, generated across 5 trials for different classifiers and probes and evaluated over 1000 test images, are plotted in Figures~\ref{fig:cifar_dropout} and \ref{fig:cifar_layers}.

Several important trends are visible in Figure~\ref{fig:cifar_dropout}.
First, interventions using any probe design improved classification accuracy from the normal performance of around 51\% to 53\%.
We emphasize that this improvement occurred despite no classifier retraining.
Second, updating representations for more gradient steps induced greater benefits.
This is visible via the different lines in the plot: results when updating for 600 steps achieved greater accuracy than when updating for 400, which outperformed only 200 steps.
Lastly, adding dropout to the probes consistently improved the effect of interventions.
Updating embeddings using probes with no dropout, as introduced in prior art (\cite{tcav,whatif}), corresponded to the leftmost data in the plot, with dropout 0.
The benefit of using dropout was most obvious when using 600 steps: model performance was further increased from 53\% accuracy (leftmost points) to 54\%.

Lastly, we investigated the effects of using dropout and multi-layer probes for interventions and plotted the results in Figure~\ref{fig:cifar_layers}.
In these experiments, all interventions were performed used 600 gradient update steps, but with probes using 1, 2, or 3 layers.
The multi-layer probes used ReLU activations for the hidden layers, allowing them to capture non-linearities in the representation space.
As before, dropout boosted performance.
Furthermore, the two-layer probe dramatically outperformed the linear probe.
Interestingly, the three-layer probe did not show as large a benefit.
Given these results, we hypothesize that using non-linear probes is beneficial (confirming findings by Tucker et al.~\shortcite{whatif}), but that there may be a limit to the benefits of larger probe models.
The TCAV baseline, which used a linear approach, corresponded to the bottom, 1-layer method.

Overall, these experiments confirmed two important trends in a classification domain:
1) using probes to inject information into representations consistently boosted performance and 2) non-linear dropout probes performed better than linear probes with no dropout.

\subsection{Avoiding Collisions in Traffic Junction}
Based on the introduction's self-driving car example, we next demonstrated how probes could be used to prevent collisions in a simulated driving scenario.
Neural agents were trained in a 2D grid setting (borrowed from prior art) to drive through an intersection without colliding, with up to 5 cars driving in the grid at the same time \cite{ic3net}.
Each agent was a feed-forward three-layer neural network mapping from inputs (current location and occupancy of the 8 neighboring grid cells) to a binary action (go/stop).
Agents received positive reward for driving through the intersection safely and 0 reward for collisions.
Over 3000 training episodes using REINFORCE, agents converged to policies in which they collided less than 1\% of the time \cite{reinforce}.

We trained 3-layer probes to predict, given the representations output by agents' second layers, whether another car was also adjacent to the intersection.
Based on the benefits of non-linear probes found in the prior section, we experimented with 2- and 3-layer probes and found no substantial difference, so we opted for the theoretically more powerful 3-layer probes.
At test time, we changed the environment to demonstrate the benefits of probe-based interventions.
We artificially corrupted one car's observations by programatically altering all observations to mask any nearby cars there.
For simplicity, we tested with only two cars at a time: one corrupted and one normal.

In this scenario, only 86\% of trials completed without a collision - where a collision was defined as the two cars entering the intersection simultaneously.
We then re-evaluated agent success rates when we used probes to update the representation of the corrupted car to reflect the true occupancy of the roads.
We performed interventions for 50 gradient updates at each timestep in the episode, evaluated over 100 episodes.

For all dropout rates from 0.0 to 0.8, we found no significant difference in the effect of interventions.
Across the 5 trained teams, the median performance when using interventions always exceeded 99\% accuracy - matching the non-corrupted environment performance.
The only exception was probes with dropout of 0.9, which achieved a success rate of 94\%.
This reinforces the pattern observed in the image classification experiments: some dropout may help, but if the dropout rate is too high, the probes fail to learn the right pattern, so interventions are not useful.

\subsection{Parent-Child}

In our final suite of experiments, we used a ``Parent-Child'' domain - a collaborative version of predator-prey from prior literature - to study the effect of probe-based interventions when humans were unaware of all information needed to produce the right action \cite{ic3net}.

\begin{figure*}[t!]
    \centering
    \begin{subfigure}[c]{0.2\textwidth}
        \centering
        \includegraphics[trim={0cm 0cm 0cm 0.0cm}, clip=true, width=0.98\textwidth]{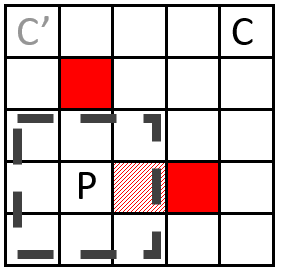}
    \end{subfigure}
    \begin{subfigure}[c]{0.39\textwidth}
        \centering
                \begin{tikzpicture}[scale=1.0] 
            \begin{axis}[legend pos=south east,  width={\textwidth},
    ylabel=Success Rate,
    xlabel=Dropout Rate,
    ymin=-0.05,ymax=1,
    xmin=-0.05, xmax=0.95,
    height=6.2cm,
    title={Interventions for Lost Child}]
            
            \addplot [black!80]
                    table [x=x, y=y, y error=y-err]{
                        x	y y-err
                        0.0	0.0 0.0
                        0.1	0.0 0.0
                        0.2	0.0 0.0
                        0.3	0.0 0.0
                        0.4	0.0 0.0
                        0.5	0.0 0.0
                        0.6	0.0 0.0
                        0.7	0.0 0.0
                        0.8	0.0 0.0
                        0.9	0.0 0.0
                            };
            \addlegendentry{Baseline}
            \addplot [mark=o, black!80, dashed, 
                      mark options={solid},
                      error bars/.cd, y dir=both, y explicit,
                      error bar style={line width=1pt,solid},
                      error mark options={line width=1pt,mark size=2pt,rotate=90}]
                    table [x=x, y=y, y error=y-err]{
                        x	y y-err
                        0.0	0.518 0.1027626391
                        0.1	0.068 0.124004129
                        0.2	0.268 0.1068317556
                        0.3	0.384 0.1131177086
                        0.4	0.53 0.1145090389
                        0.5	0.642 0.1209829079
                        0.6	0.648 0.1072089548
                        0.7	0.634 0.1072235795
                        0.8	0.730 0.1141559635
                        0.9	0.652 0.1114335677
                            };
            \addlegendentry{$t < 1$}
            
            \addplot [mark=asterisk, blue!80, dashed, 
                      mark options={solid},
                      error bars/.cd, y dir=both, y explicit,
                      error bar style={line width=1pt,solid},
                      error mark options={line width=1pt,mark size=2pt,rotate=90}]
                    table [x=x, y=y, y error=y-err]{
                        x	y y-err
                        0.0	0.582 0.09684494824
                        0.1	0.150 0.1215989474
                        0.2	0.326 0.1068901492
                        0.3	0.498 0.1145925652
                        0.4	0.638 0.1194998912
                        0.5	0.676 0.1217744801
                        0.6	0.768 0.1117511521
                        0.7	0.710 0.1108040432
                        0.8	0.776 0.1164792857
                        0.9	0.770 0.1159054787
                            };
            \addlegendentry{$t < 2$}

            \addplot [mark=square, violet!80, dashed, 
                      mark options={solid},
                      error bars/.cd, y dir=both, y explicit,
                      error bar style={line width=1pt,solid},
                      error mark options={line width=1pt,mark size=2pt,rotate=90}]
                    table [x=x, y=y, y error=y-err]{
                        x	y y-err
                        0.0	0.640 0.1013623993
                        0.1	0.284 0.115093979
                        0.2	0.450 0.10819878
                        0.3	0.736 0.1197614963
                        0.4	0.712 0.1210365895
                        0.5	0.780 0.123843385
                        0.6	0.786 0.1119982857
                        0.7	0.778 0.1132871749
                        0.8	0.790 0.1173039812
                        0.9	0.788 0.1170631283
                            };
            \addlegendentry{$t < 3$}

            \addplot [mark=o, magenta!80,
                      mark options={solid},
                      error bars/.cd, y dir=both, y explicit,
                      error bar style={line width=1pt,solid},
                      error mark options={line width=1pt,mark size=2pt,rotate=90}]
                    table [x=x, y=y, y error=y-err]{
                        x	y y-err
                        0.0	0.825 0.1037080518
                        0.1	0.692 0.1016721004
                        0.2	0.774 0.09950630131
                        0.3	0.762 0.1004872927
                        0.4	0.778 0.0993832179
                        0.5	0.800 0.1009131111
                        0.6	0.796 0.1021006562
                        0.7	0.806 0.1029542811
                        0.8	0.810 0.1107165028
                        0.9	0.804 0.108327208
                            };
            \addlegendentry{$t < 4$} 
            
            \addplot [mark=square, red!80, 
                      mark options={solid},
                      error bars/.cd, y dir=both, y explicit,
                      error bar style={line width=1pt,solid},
                      error mark options={line width=1pt,mark size=2pt,rotate=90}]
                    table [x=x, y=y, y error=y-err]{
                        x	y y-err
                        0.0	0.801 0.09481215112
                        0.1	0.786 0.08553361912
                        0.2	0.808 0.09635318365
                        0.3	0.800 0.09819784112
                        0.4	0.798 0.09763081481
                        0.5	0.800 0.0998371474
                        0.6	0.796 0.09980509005
                        0.7	0.810 0.1024816862
                        0.8	0.810 0.1080950693
                        0.9	0.804 0.1026322367
                            };
            \addlegendentry{$t < 5$}
            \end{axis} 
        \end{tikzpicture}
    \end{subfigure}
    \begin{subfigure}[c]{0.39\textwidth}
        \centering
                \begin{tikzpicture}[scale=1.0] 
            \begin{axis} [legend pos=north west, width={\textwidth},
    ylabel=Success Rate,
    xlabel=Dropout Rate,
    ymin=0.5,ymax=1,
    height=6.2cm,
    title={Interventions for Blindfolded Parent}]
            
            \addplot [black!80]
                    table [x=x, y=y, y error=y-err]{
                        x	y y-err
                        0.0	0.51 0.0
                        0.1	0.51 0.0
                        0.2	0.51 0.0
                        0.3	0.51 0.0
                        0.4	0.51 0.0
                        0.5	0.51 0.0
                        0.6	0.51 0.0
                        0.7	0.51 0.0
                        0.8	0.51 0.0
                        0.9	0.51 0.0
                            };
            \addlegendentry{Baseline}
            \addplot [mark=o, black!80, dashed, 
                      mark options={solid},
                      error bars/.cd, y dir=both, y explicit,
                      error bar style={line width=1pt,solid},
                      error mark options={line width=1pt,mark size=2pt,rotate=90}]
                    table [x=x, y=y, y error=y-err]{
                        x	y y-err
                        0.0	0.568 0.012
                        0.1	0.572 0.014
                        0.2	0.594 0.014
                        0.3	0.592 0.013
                        0.4	0.580 0.013
                        0.5	0.580 0.012
                        0.6	0.578 0.012
                        0.7	0.560 0.013
                        0.8	0.538 0.013
                        0.9	0.538 0.009
                            };
            \addlegendentry{50 steps}
            
            \addplot [mark=asterisk, blue!80, dashed, 
                      mark options={solid},
                      error bars/.cd, y dir=both, y explicit,
                      error bar style={line width=1pt,solid},
                      error mark options={line width=1pt,mark size=2pt,rotate=90}]
                    table [x=x, y=y, y error=y-err]{
                        x	y y-err
                        0.0	0.568 0.009797958971
                        0.1	0.604 0.01414213562
                        0.2	0.630 0.02244994432
                        0.3	0.666 0.0213541565
                        0.4	0.672 0.01469693846
                        0.5	0.686 0.01356465997
                        0.6	0.688 0.0228035085
                        0.7	0.676 0.02332380758
                        0.8	0.610 0.01414213562
                        0.9	0.568 0.01469693846
                            };
            \addlegendentry{100 steps}

            \addplot [mark=square, violet!80, dashed, 
                      mark options={solid},
                      error bars/.cd, y dir=both, y explicit,
                      error bar style={line width=1pt,solid},
                      error mark options={line width=1pt,mark size=2pt,rotate=90}]
                    table [x=x, y=y, y error=y-err]{
                        x	y y-err
                        0.0	0.570 0.014
                        0.1	0.612 0.019
                        0.2	0.646 0.020
                        0.3	0.676 0.022
                        0.4	0.692 0.021
                        0.5	0.734 0.018
                        0.6	0.742 0.017
                        0.7	0.744 0.020
                        0.8	0.708 0.019
                        0.9	0.618 0.008
                            };
            \addlegendentry{150 steps}

            \addplot [mark=o, magenta!80,
                      mark options={solid},
                      error bars/.cd, y dir=both, y explicit,
                      error bar style={line width=1pt,solid},
                      error mark options={line width=1pt,mark size=2pt,rotate=90}]
                    table [x=x, y=y, y error=y-err]{
                        x	y y-err
                        0.0	0.570 0.015
                        0.1	0.616 0.021
                        0.2	0.656 0.021
                        0.3	0.694 0.021
                        0.4	0.726 0.021
                        0.5	0.754 0.020
                        0.6	0.768 0.021
                        0.7	0.764 0.018
                        0.8	0.746 0.017
                        0.9	0.642 0.012
                            };
            \addlegendentry{200 steps}

            \addplot [mark=square, red!80, 
                      mark options={solid},
                      error bars/.cd, y dir=both, y explicit,
                      error bar style={line width=1pt,solid},
                      error mark options={line width=1pt,mark size=2pt,rotate=90}]
                    table [x=x, y=y, y error=y-err]{
                        x	y y-err
                        0.0	0.572 0.016
                        0.1	0.620 0.022
                        0.2	0.658 0.022
                        0.3	0.700 0.023
                        0.4	0.746 0.022
                        0.5	0.774 0.023
                        0.6	0.790 0.020
                        0.7	0.848 0.018
                        0.8	0.798 0.016
                        0.9	0.690 0.014
                            };
            \addlegendentry{250 steps}
            \end{axis} 
        \end{tikzpicture}
    \end{subfigure}
    \caption{In the parent-child environment, the blindfolded parent (P) did not observe nearby obstacles, and the lost child (C) observed its location as $(0, 0)$ even when located at $(0, 4)$. In ``lost-child'' settings (middle), interventions caused the parent to find the child with nearly 80\% accuracy, whereas without interventions the success rate was 0\%. In the ``blindfolded-parent'' setting, interventions similarly boosted performance. Especially with the blindfolded-parent, probes with dropout between 0.5 and 0.8 often outperformed probes without dropout.}
    \label{fig:parentchild}
\end{figure*}
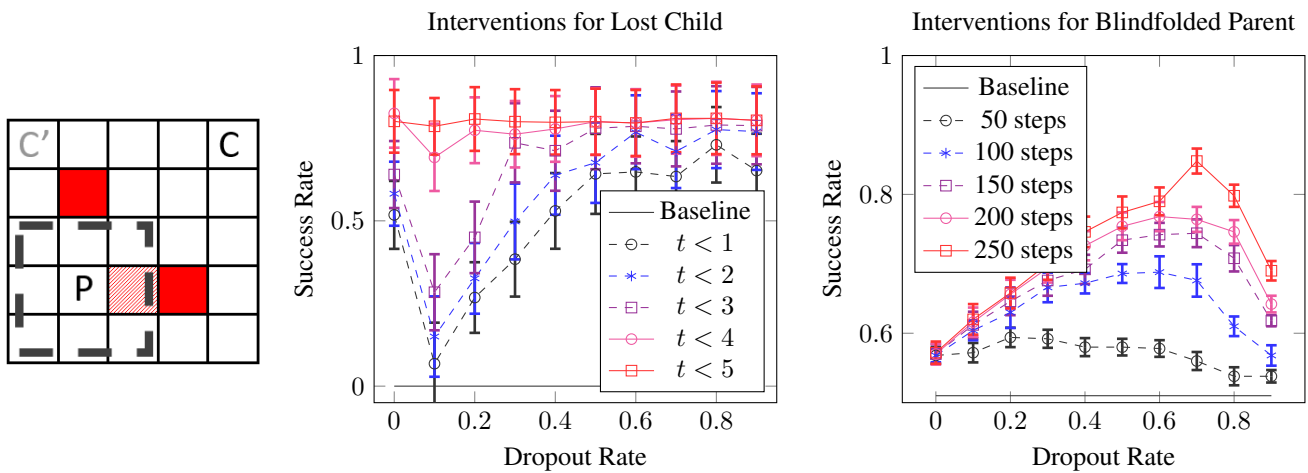

In this two-agent setting, a stationary child and a movable parent were spawned in random locations in a $5\times5$ grid.
As depicted in Figure~\ref{fig:parentchild}, the parent observed its location and the surround cell locations at all timesteps; the child only observed its location at timestep 0.
At the start of each episode, each grid location was populated by an obstacle with 10\% probability.
The agents received shared reward of 1 if the parent ended the episode (after 20 timesteps) at the child's location and otherwise got reward 0.
If the parent collided with an obstacle, it could not move for the rest of the episode.

Agents were instantiated using the LSTM-based IC3Net architecture, allowing them to communicate with each other agent at all timesteps, and trained with REINFORCE \cite{ic3net,reinforce}.
The parent could move up, down, left, right, or stay still; the child could not move.
Thus, the optimal policy was for the child to communicate its location to the parent and for the parent to navigate to the child, avoiding obstacles along the way.

We trained 5 parent-child teams to convergence using different random seeds to convergence at 100\% success rates.
For a given team, for both the parent and child, we trained $h$ and $c$ probes that accepted the LSTM's hidden or cell states, respectively.
The parent probes were trained to predict the obstacle occupancy grid of the cells surrounding the parent; the child probes were trained to predict the child's location in that 2D world.
To gain further statistical power, for each parent-child team, we trained probes from scratch 5 times with different random seeds.
Overall, given the 5 pairs of agents, 2 types of probes per agent, 5 random seeds for probes, and 10 dropout rate levels, this corresponded to 500 trained probes.

As in the traffic junction experiment, we created ``corrupted'' versions of these environments to see if probes could be used to overcome perceptual failures of the agents.
For consistency, all corrupted versions started with the same initial conditions: the child in the upper right and the parent in the lower left.
In corrupting the child's perception, we replaced the observation of its true location with data from the upper left.
In corrupting the parent's perception, we masked all observations of obstacles.
We then tested if, using our probes, we could remedy child corrupted data alone, parent corrupted data alone, or even both simultaneously.
We used both the $h$ and $c$ probes to update both representations when performing interventions.

Our results in the lost-child settings are reported in Figures~\ref{fig:parentchild}.
In this experiment, wherein the child was actually at $(4, 0)$ but observed its location as $(0, 0)$, the baseline performance without interventions was 0\%: that is, the parent never found the child.
Conversely, using interventions, we increased the success rate to up to 80\%, as shown in Figure~\ref{fig:parentchild}.
On the $x$ axis, we varied the dropout rates of the probes, on the $y$ axis we measured the team success rates.
The different curves correspond to intervening on the child's representations for the first timestep, first and second timestep, etc..
For all probe types, intervening for more timesteps always helped: however, intervening for only the first 2 timesteps with probes with dropout rate 0.9 was still enough to have the parent reach the child nearly 80\% of the time.

Performance as a function of dropout was more subtle, though: small but positive dropout rates seemed to worsen behavior when only performing interventions for one or two timesteps, but as the dropout rate increased above 0.5, interventions became more effective than when not using dropout.
This non-linear behavior is surprising and warrants further investigation in future work.

Similarly, we measured success rate as a function of probe dropout rate in the blindfolded-parent environment.
Without interventions, because the parent could not observe obstacles, roughly 50\% of episodes resulted in collisions and therefore failures.
Conversely, when we intervened to update the parent representations, especially when using dropout probes, we decreased the likelihood of collisions and increased the success rate.
In addition to varying our analysis by dropout rate, we also investigated the effect of more gradient steps when updating representations and plotted the results as different curves.
For all dropout rates, more gradient steps produced better effects, although the effect was greatest for dropout rates between 0.1 and 0.8.

Lastly, in experiments conducted in which both the parent and child observations were corrupted, we demonstrated that probe-based interventions on both agents mitigated the observation failures, but that higher-dropout probes were most effective.
Success rate without interventions was 0\%, but reached 72\% for probes with dropout 0.7.
Full results are omitted for brevity but are available in Appendix~\ref{app:both}.

\section{Discussion}
Jointly, the results from the prior section demonstrate important benefits of test-time interventions on representations.
First, we showed how an imperfect image classifier could be improved by injecting high-level information into representations.
This indicates a possible strategy for companies seeking to improve AI performance without retraining a model from scratch.
Second, in the traffic junction experiment, we demonstrated that, in simple cases where the injected information and expected behavior had a clear relationship, we could induce the right behavior.
Although simplified, this demonstrates how humans might be able to affect the behavior of AI systems that they cannot directly control.

Lastly, in the parent-child environment, we showed more subtle capabilities.
Despite not understanding the emergent communication protocol, humans were able to interact with agents by only updating the relevant parts of representations.
This therefore illustrates the value of our method in cases in which humans may want to guide agent behavior without dictating all aspects of it.
For example, in the blidfolded-parent scenario, the human could indicate the presence of obstacles that should be avoided without having to know what the goal location was (and therefore whether the parent should avoid an obstacle by going above or below it).


\section{Conclusion}
In this work, we proposed probe-based interventions as a mechanism for inserting information into neural model representations - a potential approach to the human-assisted decision-making problem.
We found that prior methods, initially developed for model explanability work could be adapted to this problem, and we showed how a new, dropout-based intervention method often outperformed other techniques.
We tested our method in a variety of contexts, from image classification to emergent communication domains, demonstrating the wide applicability of our approach.
While we have taken an initial step towards supporting human interventions, future work could develop more sophisticated intervention methods or evaluate the tradeoffs between minimal and sufficient changes to representations.
Lastly, we are particularly interested in continuing work to train models that better respond to interventions.


\bibliographystyle{named}
\bibliography{ijcai22}

\begin{thebibliography}{}

\bibitem[\protect\citeauthoryear{Alain and
  Bengio}{2016}]{alain2016understanding}
Guillaume Alain and Yoshua Bengio.
\newblock Understanding intermediate layers using linear classifier probes.
\newblock {\em arXiv preprint arXiv:1610.01644}, 2016.

\bibitem[\protect\citeauthoryear{Belghazi \bgroup \em et al.\egroup
  }{2018}]{mine}
Mohamed~Ishmael Belghazi, Aristide Baratin, Sai Rajeshwar, Sherjil Ozair,
  Yoshua Bengio, Aaron Courville, and Devon Hjelm.
\newblock Mutual information neural estimation.
\newblock In Jennifer Dy and Andreas Krause, editors, {\em Proceedings of the
  35th International Conference on Machine Learning}, volume~80 of {\em
  Proceedings of Machine Learning Research}, pages 531--540. PMLR, 10--15 Jul
  2018.

\bibitem[\protect\citeauthoryear{Bernstein \bgroup \em et al.\egroup
  }{2013}]{bernstein2013complexity}
Daniel~S Bernstein, Shlomo Zilberstein, and Neil Immerman.
\newblock The complexity of decentralized control of markov decision processes,
  2013.

\bibitem[\protect\citeauthoryear{Chen \bgroup \em et al.\egroup
  }{2020}]{chen2020concept}
Zhi Chen, Yijie Bei, and Cynthia Rudin.
\newblock Concept whitening for interpretable image recognition.
\newblock {\em Nature Machine Intelligence}, 2(12):772--782, 2020.

\bibitem[\protect\citeauthoryear{{Elazar} \bgroup \em et al.\egroup
  }{2020}]{elazar2020amnesic}
Yanai {Elazar}, Shauli {Ravfogel}, Alon {Jacovi}, and Yoav {Goldberg}.
\newblock {Amnesic Probing: Behavioral Explanation with Amnesic
  Counterfactuals}.
\newblock {\em arXiv e-prints}, page arXiv:2006.00995, June 2020.

\bibitem[\protect\citeauthoryear{Giulianelli \bgroup \em et al.\egroup
  }{2018}]{diagnostic}
Mario Giulianelli, Jack Harding, Florian Mohnert, Dieuwke Hupkes, and Willem
  Zuidema.
\newblock Under the hood: Using diagnostic classifiers to investigate and
  improve how language models track agreement information.
\newblock In {\em Proceedings of the 2018 {EMNLP} Workshop {B}lackbox{NLP}:
  Analyzing and Interpreting Neural Networks for {NLP}}, pages 240--248,
  Brussels, Belgium, November 2018. Association for Computational Linguistics.

\bibitem[\protect\citeauthoryear{John \bgroup \em et al.\egroup
  }{2019}]{john-etal-2019-disentangled}
Vineet John, Lili Mou, Hareesh Bahuleyan, and Olga Vechtomova.
\newblock Disentangled representation learning for non-parallel text style
  transfer.
\newblock In {\em Proceedings of the 57th Annual Meeting of the Association for
  Computational Linguistics}, pages 424--434, Florence, Italy, July 2019.
  Association for Computational Linguistics.

\bibitem[\protect\citeauthoryear{Kim \bgroup \em et al.\egroup }{2018}]{tcav}
Been Kim, Martin Wattenberg, Justin Gilmer, Carrie Cai, James Wexler, Fernanda
  Viegas, et~al.
\newblock Interpretability beyond feature attribution: Quantitative testing
  with concept activation vectors (tcav).
\newblock In {\em International conference on machine learning}, pages
  2668--2677. PMLR, 2018.

\bibitem[\protect\citeauthoryear{Krizhevsky \bgroup \em et al.\egroup
  }{}]{cifar10}
Alex Krizhevsky, Vinod Nair, and Geoffrey Hinton.
\newblock Cifar-10 (canadian institute for advanced research).

\bibitem[\protect\citeauthoryear{Molnar}{2019}]{molnar2019}
Christoph Molnar.
\newblock {\em Interpretable Machine Learning}.
\newblock 2019.

\bibitem[\protect\citeauthoryear{PyTorch}{2022}]{classifier}
PyTorch.
\newblock Training a classifier, 2022.

\bibitem[\protect\citeauthoryear{Ribeiro \bgroup \em et al.\egroup
  }{2016}]{ribeiro2016model}
Marco~Tulio Ribeiro, Sameer Singh, and Carlos Guestrin.
\newblock Model-agnostic interpretability of machine learning.
\newblock {\em arXiv preprint arXiv:1606.05386}, 2016.

\bibitem[\protect\citeauthoryear{Singh \bgroup \em et al.\egroup
  }{2018}]{ic3net}
Amanpreet Singh, Tushar Jain, and Sainbayar Sukhbaatar.
\newblock Learning when to communicate at scale in multiagent cooperative and
  competitive tasks.
\newblock {\em arXiv preprint arXiv:1812.09755}, 2018.

\bibitem[\protect\citeauthoryear{Tucker \bgroup \em et al.\egroup
  }{2021}]{whatif}
Mycal Tucker, Peng Qian, and Roger Levy.
\newblock What if this modified that? syntactic interventions with
  counterfactual embeddings.
\newblock In {\em Findings of the Association for Computational Linguistics:
  ACL-IJCNLP 2021}, pages 862--875, Online, August 2021. Association for
  Computational Linguistics.

\bibitem[\protect\citeauthoryear{Williams}{1992}]{reinforce}
Ronald~J Williams.
\newblock Simple statistical gradient-following algorithms for connectionist
  reinforcement learning.
\newblock {\em Machine learning}, 8(3):229--256, 1992.

\bibitem[\protect\citeauthoryear{Zeiler and
  Fergus}{2014}]{zeiler2014visualizing}
Matthew~D Zeiler and Rob Fergus.
\newblock Visualizing and understanding convolutional networks.
\newblock In {\em European conference on computer vision}, pages 818--833.
  Springer, 2014.

\end{thebibliography}

\newpage
\appendix
\section{Redundancy Analysis}
\label{app:mine}
The main paper motivates dropout probes from a theory of redundant representations; here, we use a technique from prior art to demonstrate that such redundancy existed in one of our experiments.

We define a measure of property-specific representational redundancy as follows:

\begin{equation}
    R(S, Z, Z_1, Z_2) = I(S, Z_1) + I(S, Z_2) - I(S, Z)
\end{equation}

That is, redundancy, $R$, in nats, is equal to the difference of two mutual information terms.
The first, $I(S, Z_1) + I(S, Z_2)$, equals the sum of the mutual information between $Z_1$ and $S$ and $Z_2$ and $S$ where we let $S$ be a random variable describing a property of interest, and $Z_1$ and $Z_2$ are mutually-exclusive and collectively exhaustive subsets of a $Z$.
The second term is merely the mutual information between a representation, $Z$, and $S$.

We applied this metric to study parents in the parent-child domain.
$Z$ was defined as the 128-dimensional hidden state; $Z_1$ and $Z_2$ were the first and second halves of $Z$, respectively.
Lastly, $S$ was the 8-dimensional binary vector representing the presence of obstacles in the cells surrounding the parent.
In other words, we sought to measure how much ``the information about nearby obstacles that is in the whole representation is copied in the first and second halves.''

Analytically calculating the terms for computing $R$ is challenging, but a technique dubbed MINE may be used to quickly approximate pairwise mutual information \cite{mine}.
In essence, the technique trains a neural net to maximize a lower bound on the mutual information between pairs of variables.

We therefore employed MINE to calculate the 3 terms necessary for measuring the redundancy of obstacle information in parents' hidden states.
Averaged over the 5 trained teams, $I(S, Z_1) \approx I(S, Z_2) = 0.7$ and $I(S, Z) =0.8$.
(All standard deviations were less than 0.1.)
Therefore, as measured in nats, $R(S, Z, Z_1, Z_2) = 0.7 + 0.7 - 0.8 = 0.6$.
That is, there are 0.6 nats of common information about the presence of obstacles in the first and second halves of $Z$.
This proves that the information is encoded redundantly, which in turn motivates dropout probes for interventions.

We repeated such experiments for the latent representations used in the traffic experiment (for $S$ denoting if another car was near the intersection) and for the hidden and cell state activations for both parent and child in the parent-child settings (for $S$ denoting nearby obstacles for parents and the child's location for children).
In all cases, we found positive redundancy, indicating that this phenomenon is widespread.

Here, we have only shown that redundancy is positive given our choice of $Z_1$ and $Z_2$ as the first and second halves of a representation.
This suffices for a proof by existence, but other work defining different subsets of representations could better characterize redundancy.

\section{Parent-Child Full Results}
\label{app:both}
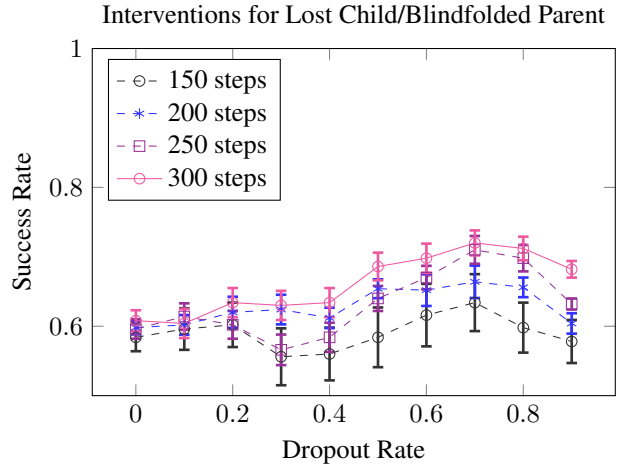
\begin{figure}[t!]
    \centering
    \begin{subfigure}[c]{0.48\textwidth}
        \centering
                \begin{tikzpicture}[scale=1.0] 
            \begin{axis} [legend pos=north west, width={\textwidth},
    ylabel=Success Rate,
    xlabel=Dropout Rate,
    ymin=0.5,ymax=1,
    height=6.2cm,
    title={Interventions for Lost Child/Blindfolded Parent}]
            
            \addplot [mark=o, black!80, dashed, 
                      mark options={solid},
                      error bars/.cd, y dir=both, y explicit,
                      error bar style={line width=1pt,solid},
                      error mark options={line width=1pt,mark size=2pt,rotate=90}]
                    table [x=x, y=y, y error=y-err]{
                        x	y y-err
                        0.0	0.584 0.02
                        0.1	0.596 0.03
                        0.2	0.602 0.032
                        0.3	0.556 0.041
                        0.4	0.560 0.038
                        0.5	0.584 0.043
                        0.6	0.616 0.045
                        0.7	0.634 0.041
                        0.8	0.598 0.036
                        0.9	0.578 0.031
                            };
            \addlegendentry{150 steps}
            
            \addplot [mark=asterisk, blue!80, dashed, 
                      mark options={solid},
                      error bars/.cd, y dir=both, y explicit,
                      error bar style={line width=1pt,solid},
                      error mark options={line width=1pt,mark size=2pt,rotate=90}]
                    table [x=x, y=y, y error=y-err]{
                        x	y y-err
                        0.0	0.598 0.009797958971
                        0.1	0.602 0.01414213562
                        0.2	0.620 0.02244994432
                        0.3	0.624 0.0213541565
                        0.4	0.612 0.01469693846
                        0.5	0.654 0.01356465997
                        0.6	0.652 0.0228035085
                        0.7	0.664 0.02332380758
                        0.8	0.656 .01414213562
                        0.9	0.604 0.01469693846
                            };
            \addlegendentry{200 steps}

            \addplot [mark=square, violet!80, dashed, 
                      mark options={solid},
                      error bars/.cd, y dir=both, y explicit,
                      error bar style={line width=1pt,solid},
                      error mark options={line width=1pt,mark size=2pt,rotate=90}]
                    table [x=x, y=y, y error=y-err]{
                        x	y y-err
                        0.0	0.596 0.014
                        0.1	0.614 0.019
                        0.2	0.602 0.020
                        0.3	0.566 0.022
                        0.4	0.584 0.021
                        0.5	0.640 0.018
                        0.6	0.670 0.017
                        0.7	0.710 0.020
                        0.8	0.698 0.019
                        0.9	0.632 0.008
                            };
            \addlegendentry{250 steps}

            \addplot [mark=o, magenta!80,
                      mark options={solid},
                      error bars/.cd, y dir=both, y explicit,
                      error bar style={line width=1pt,solid},
                      error mark options={line width=1pt,mark size=2pt,rotate=90}]
                    table [x=x, y=y, y error=y-err]{
                        x	y y-err
                        0.0	0.608 0.015
                        0.1	0.604 0.021
                        0.2	0.634 0.021
                        0.3	0.630 0.021
                        0.4	0.634 0.021
                        0.5	0.686 0.020
                        0.6	0.698 0.021
                        0.7	0.720 0.018
                        0.8	0.712 0.017
                        0.9	0.682 0.012
                            };
            \addlegendentry{300 steps} 
            \end{axis} 
        \end{tikzpicture}
    \end{subfigure}
    \caption{When both the child and parent observations were corrupted, probe-based inerventions improved over baseline performance, particularly when using higher-dropout probes ($x$ axis) for more gradient steps (different curves).}
    \label{fig:parentchild_app}
\end{figure}

In the main paper, we omitted the full results of experiments in which both the parent and child observations were corrupted.
Here, we include the full results in Figure~\ref{fig:parentchild_app}.

The baseline success rate, without interventions, was zero.
However, with interventions, the parent found the child up to 72\% of the time.
This indicates that the high-dropout probes were able to update both the child's representation of its location and the parent's representation of nearby obstacles.
This hints at a sort of compositionality of interventions.

\section{Implementation Details}
The CIFAR10 classifier was trained per the hyperparameters specified in the online tutorial \cite{cifar10,classifier}.

The traffic agents were trained using the IC3Net architecture for feedfoward agents \cite{ic3net}.
Each agent comprised 4 dense layers with hidden dimension 128; weights were shared among agents.
Agents were trained with an RMSProp optimizer with learning rate 0.001, $\alpha = 0.97$, and $\epsilon = 10^{-6}$, the default values provided with IC3Net, for 3000 episodes.

The parent-child agents were trained using the IC3Net architecture for LSTM-based agents \cite{ic3net}.
Agents shared weights and were parametrized with 2 dense layers encoding observations with hidden dimension 128 and an LSTM cell with 128-dimensional $c$ and $h$ cells.
Agents were trained for 1000 episodes, with the same optimizer as the one for the traffic experiments.

\end{document}